\documentclass[lettersize,journal]{IEEEtran}
\usepackage{amssymb}
\usepackage{amsthm}
\usepackage[utf8]{inputenc}
\usepackage{amsmath}
\usepackage{graphicx} 
\usepackage{orcidlink}
\usepackage{multirow}
\usepackage{graphics}
\usepackage{graphicx}
\usepackage{booktabs}
\usepackage{setspace}
\usepackage{array}
\usepackage{booktabs}
\usepackage{cite}
\usepackage{multirow}
\usepackage{subcaption}
\usepackage{tabularx}
\usepackage{balance}

\usepackage{makecell}
\usepackage{tikz,xcolor,hyperref}
\usepackage[caption=false,font=normalsize,labelfont=sf,textfont=sf]{subfig}
\usepackage{bm}

\usepackage{calc}
\usepackage{pgf}
\usepackage[switch]{lineno}
\usepackage{algorithm}
\usepackage{algorithmic}

\newcommand{\orcidauthorA}{0009-0002-9031-3176}
\newcommand{\orcidauthorB}{0000-0001-7303-5712}
\newcommand{\orcidauthorC}{0000-0003-4352-1431}
\newcommand{\orcidauthorD}{0000-0002-8581-9554}
\newcommand{\orcidauthorE}{0000-0002-2812-8781}
\makeatletter
\let\sf@counterlist\@empty
\makeatother
\begin{document}

\title{AvatarMakeup: Realistic Makeup Transfer for 3D Animatable Head Avatars}

\author{
    Yiming Zhong$^{\orcidlink{\orcidauthorA}}$, Xiaolin Zhang$^\dagger$$^{\orcidlink{\orcidauthorB}}$, Ligang Liu$^{\orcidlink{\orcidauthorC}}$, Yao Zhao$^{\orcidlink{\orcidauthorD}}$, and Yunchao Wei$^\dagger$$^{\orcidlink{\orcidauthorE}}$
\thanks{Corresponding author: Xiaolin Zhang, Yunchao Wei.}
\thanks{Yiming Zhong, Yunchao Wei and Yao zhao are with the Institute of Information Science and Visual Intelligence + X International Joint Laboratory, Beijing Jiaotong University, Beijing, China. (e-mail: ymzhong@bjtu.edu.cn,  wychao1987@gmail.com, yzhao@bjtu.edu.cn).}
\thanks{Xiaolin Zhang is with the College of Electrical Engineering and Automation, Shandong University of Science and Technology. (e-mail: solli.zhang@gmail.com).}
\thanks{Ligang Liu is with the School of Mathematical Sciences, University of Science and Technology of China, Hefei. (e-mail: lgliu@ustc.edu.cn).}
}



\maketitle

\begin{abstract}
Similar to facial beautification in real life, 3D virtual avatars require personalized customization to enhance their visual appeal, yet this area remains insufficiently explored.
Although current 3D Gaussian editing methods can be adapted for facial makeup purposes, these methods fail to meet the fundamental requirements for achieving realistic makeup effects: 1) ensuring a consistent appearance during drivable expressions, 2) preserving the identity throughout the makeup process, and 3) enabling precise control over fine details.
To address these, we propose a specialized 3D makeup method named \textit{AvatarMakeup}, leveraging a pretrained diffusion model to transfer makeup patterns from a single reference photo of any individual.
We adopt a \textit{coarse-to-fine} idea to first maintain the consistent appearance and identity, and then to refine the details. 
In particular, the diffusion model is employed to generate makeup images as supervision. Due to the uncertainties in diffusion process, the generated images are inconsistent across different viewpoints and expressions. 
Therefore, we propose a \textit{Coherent Duplication} method to coarsely apply makeup to the target while ensuring consistency across dynamic and multiview effects. 
Coherent Duplication optimizes a global UV map by recoding the averaged facial attributes among the generated makeup images.
By querying the global UV map, it easily synthesizes coherent makeup guidance from arbitrary views and expressions to optimize the target avatar.
Given the coarse makeup avatar, we further enhance the makeup by incorporating a \textit{Refinement Module} into the diffusion model to achieve high makeup quality.  
Experiments demonstrate that AvatarMakeup achieves state-of-the-art makeup transfer quality and consistency throughout animation.
\end{abstract}


\newcommand{\etal}{\textit{et~al.\,}}
\newcommand{\eg}{\textit{e.g.}}
\newcommand{\ie}{\textit{i.e.}}
\newcommand{\wrt}{\textit{w.r.t.}}
\newcommand{\etc}{\textit{etc.}}
\newcommand{\red}{\textcolor[rgb]{1,0,0}}
\newcommand{\green}{\textcolor[rgb]{0,1,0}k k}
\newcommand{\blue}{\textcolor[rgb]{0,0,1}}


\begin{figure*}[ht]
  \centering
    \includegraphics[width=1\linewidth,]{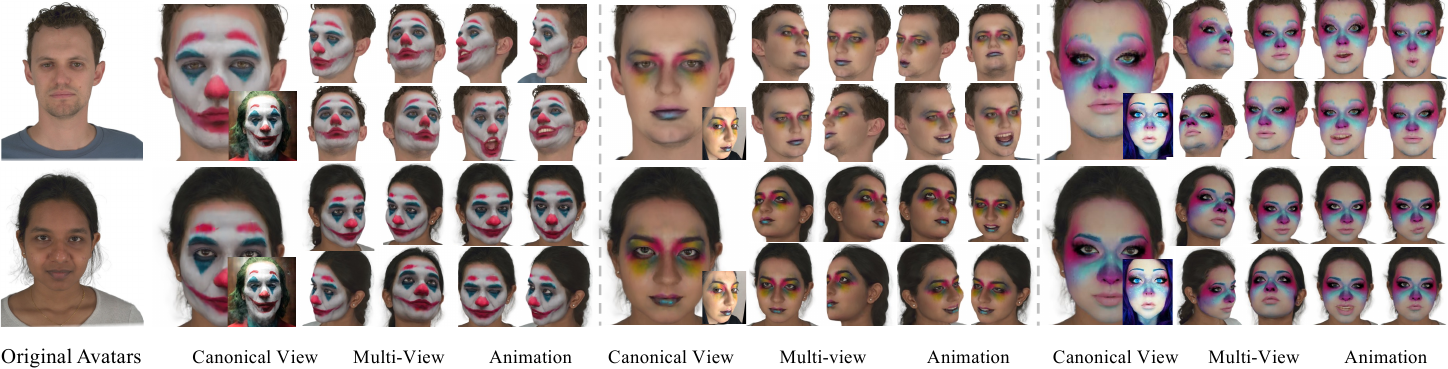}
  \caption{3D makeup transfer examples generated by AvatarMakeup. We improve the quality of makeup transfer by employing a coarse-to-fine strategy.  Examples show that under multi-view and animation conditions, our method generates high-quality and consistent makeup effects while maintaining the identity.}
  \label{fig:display}
\end{figure*}
\section{Introduction}
\label{sec:intro}

\IEEEPARstart{R}{ecently}, 3D representations using Gaussian Splatting~\cite{kerbl20233d}(3DGS) have attracted significant attention for their highly realistic rendering quality and remarkable real-time efficiency. Researchers have developed animatable 3D avatar models~\cite{Qian_2024_CVPR, shao2024splattingavatar} based on Gaussian Splatting. These methods enable dynamic, lifelike character animations with high fidelity, facilitating applications in virtual reality, gaming, and immersive environments. Like real-world preferences, users in 3D avatar applications increasingly seek beautification and makeup customization options to enhance and personalize their virtual presence.



Existing models~\cite{khan2025instaface,li2018beautygan, jiang2020psgan, deng2021spatially, nguyen2021lipstick, xiang2022ramgan, gu2019ladn, liu2021psgan++, wan2022facial, yan2023beautyrec, sun2023ssat, kips2020gan, yang2022elegant, zhang2024stablemakeuprealworldmakeuptransfer} have achieved considerable success in facial beautification and editing within 2D avatars.
For example, Generative Adversarial Network (GAN)-based approaches~\cite{li2018beautygan, jiang2020psgan, deng2021spatially, nguyen2021lipstick, xiang2022ramgan, gu2019ladn, liu2021psgan++, wan2022facial, yan2023beautyrec, sun2023ssat, kips2020gan, yang2022elegant} demonstrate high robustness and generalizability across various makeup styles. Stable-Makeup~\cite{zhang2024stablemakeuprealworldmakeuptransfer} achieves high fidelity makeup transfer.  It constructs a comprehensive dataset encompassing diverse makeup styles and finetunes a pretrained diffusion model. 

However, these models are limited to facial editing within 2D images due to the lack of paired 3D makeup datasets.  Fully extending the facial makeup application of 3D avatars remains challenging. 
An attemptable approach to address this task is to utilize the previous 3D Gaussian editing methods.
Particularly, Geneavatar~\cite{Bao_2024_CVPR} generates consistent makeup information by 3DMM-based 3DGAN~\cite{sun2023next3d} and subsequently optimizes a NeRF-represented avatar. Nevertheless, the GAN generator struggles to fit intricate and creative makeup details, and Geneavatar also falls short in achieving real-time rendering. 
GaussianEditor~\cite{chen2024gaussianeditor}, DGE Editor~\cite{chen2024dge} and TIP-Editor~\cite{zhuang2024tip} proposed for the representation of Gaussian Splatting~\cite{kerbl20233d} have made strides in editing 3D Gaussian objects and scenes by leveraging textual instructions to guide modifications. Unfortunately, these methods have two key limitations for 3D facial makeup: 1) These methods are limited to editing static representations and cannot achieve the dynamic makeup effects required for animatable human faces. 2) The primary objective of facial makeup transfer is to preserve the identity of the target character, yet these methods fail to account for this crucial aspect. 

Therefore, we conduct makeup transfer by addressing the limiatations. We believe that makeup transfer for 3D avatars should meet two fundamental requirements: 1) Facial makeup should be extended to be applied on rigged avatars for animation purpose; 2) Facial makeup requires precise control over the details to achieve beautiful and refined looks while preserving the identity of the original individuals.
%
In this paper, we present a novel framework named \textit{AvatarMakeup} to execute makeup transfer for rigged 3D Gaussian avatars from 2D makeup methods. To make up animatable avatars, our method inherits the animation module from recent works on reconstructing rigged gaussian avatars~\cite{Qian_2024_CVPR, shao2024splattingavatar}. Specifically, those works establish binding connections between 3D Gaussians and FLAME mesh~\cite{FLAME:SiggraphAsia2017} to make 3D gaussian kernels uniformly distributed over the surface of the mesh. Therefore, 3D gaussian avatars can be animated by adjusting the FLAME parameters. To precisely control the makeup details, unlike previous methods~\cite{chen2024gaussianeditor} using textual descriptions to edit facial makeup, our methods derived makeover details from a reference image from any person. We believe that facial editing guided by image-based conditioning offers a more refined and natural approach compared to language-based conditioning. 
 
Intuitively, we adopt a coarse-to-fine strategy to first maintain consistent appearance and identity and then refine the details. The strategy intuitively imitates the process akin to how a human would apply makeup. 
The process begins with applying base makeup and then delicate makeup. We leverage Stable-Makeup to transfer makeup patterns from a single reference photo of any individual.
In practice, Stable-Makeup generates novel-view and various expression makeup images as supervision.
This supervision information is employed to guide the makeup process of 3D avatars. 
Due to the inherent uncertainty in the diffusion process, the images generated by Stable-Makeup often exhibit inconsistencies, resulting in artifacts when driving avatars with extreme poses and expressions. 
To address this, we propose a novel \textit{Coherent Duplication} method that coarsely applies makeup to the target while maintaining consistency across dynamic and multiview effects. 
In detail, given the generated images, our method utilizes the bonded mesh to create a global UV map, which captures and records the basic facial patterns. This enables a consistent representation of facial features across various poses and expressions, ensuring more coherent and accurate makeup application. By querying the constructed UV map, Coherent Duplication synthesizes coarse yet consistent makeup images from novel viewpoints and expressions with ease. These images serve as supervision to optimize the Gaussian avatars, effectively balancing quality and consistency during animation.

Building upon the coarse makeup, we further propose a \textit{Refinement Module} into the 3D makeup process to enrich the avatars with intricate makeup details. Specifically, we introduce noise with a small timestamp during the diffusion process. This approach not only eliminates blurred details but also ensures the consistency of the base makeup. As a result, the optimized avatars achieve high-quality makeup while maintaining consistency throughout animation. The outcomes of the proposed AvatarMakeup method are demonstrated in Fig~\ref{fig:display}. 

In summary, our contributions are as follows: 
\begin{itemize}

\item This paper proposes AvatarMakeup, a novel framework to apply makeover transfer to animatable head avatars. The method precisely transfers makeup styles from any person to the target avatars. 

\item We present a Coherent Duplication method that utilizes the mesh bonded to 3D gaussians to provide consistent makeover information across diverse viewpoints and expressions.

\item Experimental results show that our AvatarMakeup achieves state-of-the-art performance, reflected in the transferring quality and multi-view consistency.

\end{itemize}

\begin{figure*}[ht]
  \centering
  \includegraphics[width=0.95\linewidth,]{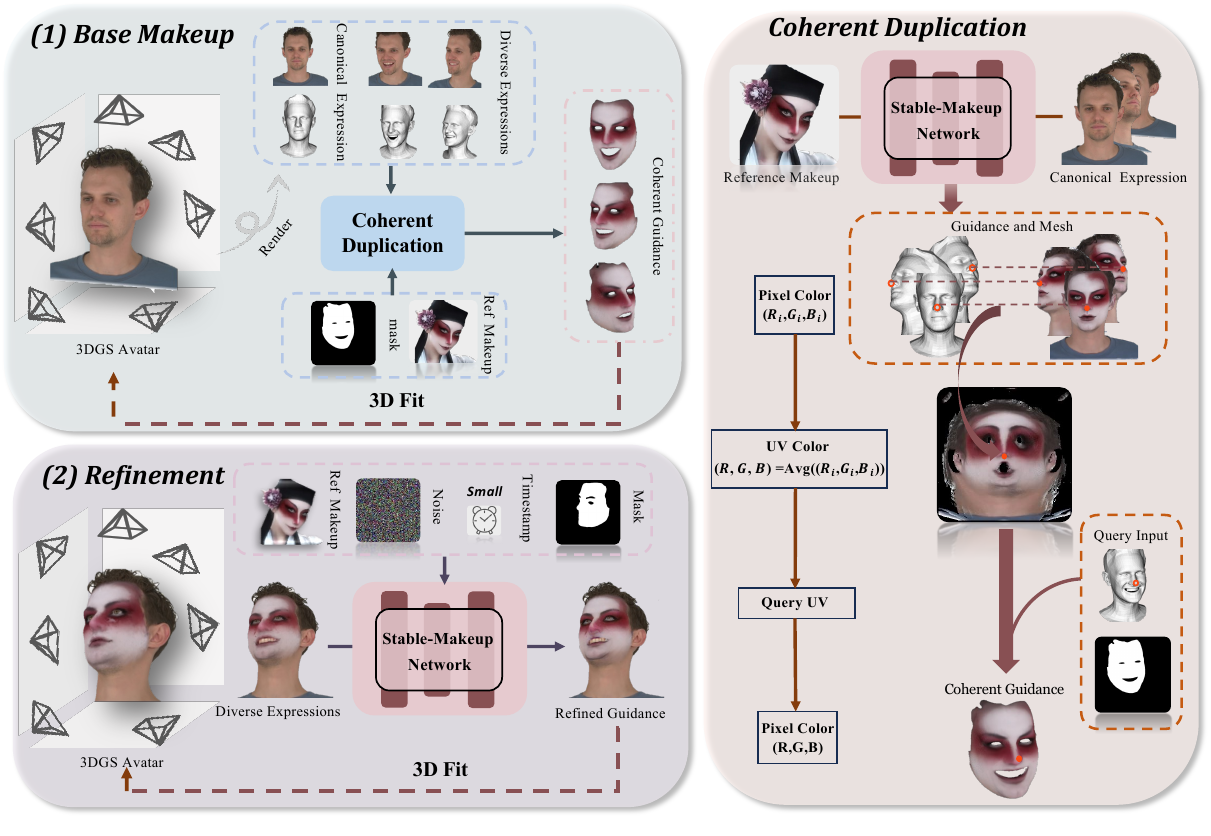}
  \caption{Illustration of AvatarMakeup. AvatarMakeup takes a reconstructed avatar and a reference makeup image as input and employs a coarse-to-fine pipeline to gradually apply the makeup to the target avatar. (1) In the coarse stage, we propose Coherent Duplication methods to generate consistent guidance images. (2) In the refinement stage, AvatarMakeup refines the base makeup by integrating a refinement strategy into the Stable-Makeup model. (3) The Coherent Duplication method uses FLAME mesh to construct a global UV map. By querying the UV map, we can easily generate coherent guidance images from arbitrary views and expressions.
  }
  \label{fig:pipeline}
\end{figure*}
\section{Related works}
\label{sec:related}

\subsection{3D Animatable Avatars}
The advancement of animatable avatar reconstruction primarily relies on the progress made in different representation, 
with parametric frameworks like SMPL~\cite{loper2023smpl} and FLAME~\cite{FLAME:SiggraphAsia2017} serving as foundational tools. Face2face ~\cite{thies2016face2face} pioneers the direction toward digital avatars through real-time facial tracking and realistic face reenactment. Then many methods use mesh to represent the avatars in 3D space. PIFu~\cite{saito2019pifu}and PIFuHD\cite{saito2020pifuhd} introduce pixel-aligned implicit functions to reconstruct clothed humans from single images. ARCH~\cite{huang2020arch} and ARCH++~\cite{he2021arch++} extend this by incorporating animatable parametric models, enabling pose-aware reconstruction of clothed avatars. For head avatars,  HiFace~\cite{chai2023hiface} disentangles static and dynamic facial details for high-fidelity reconstruction, while Vid2Avatar~\cite{guo2023vid2avatar} reconstructs animatable head avatars from monocular video via neural rendering.
Neural Radiance Field (NeRF)~\cite{mildenhall2021nerf} restores the avatars's information implicitly and enables capturing high-frequency avatar details. HumanNeRF~\cite{isik2023humanrf} first to extend NeRF to dynamic humans using SMPL-guided deformation fields, enabling free-viewpoint rendering of moving subjects from monocular video. InstantAvatar~\cite{jiang2023instantavatar} accelerates training via hash encoding while maintaining animatable properties through learned deformation fields. 
Gafni et al. ~\cite{gafni2021dynamic} developed a NeRF conditioned on an expression vector from monocular videos. Grassal et al.~\cite{Grassal_2022_CVPR} enhanced FLAME by subdividing it and adding offsets to improve its geometry, allowing for a dynamic texture created by an expression-dependent texture field. IMavatar~\cite{zheng2022avatar} constructs a 3D animatable head avatar utilizing neural implicit functions, creating a mapping from observed space to canonical space through iterative root-finding. HeadNeRF\cite{hong2022headnerf} implements a NeRF-based parametric head model incorporating 2D neural rendering for improved efficiency. INSTA~\cite{zielonka2023instant} deforms query points to a canonical space by finding the nearest triangle on a FLAME mesh and combining this with InstantNGP~\cite{muller2022instant} to achieve fast rendering. After 3D Gaussian Splatting(3DGS)~\cite{kerbl20233d} occurred,  the representation benefits avatar reconstruction with real-time rendering and fine-grained details. On the one hand, many methods animate avatars by decoding facial latents to 3D Gaussians based on animation parameters. HeadGas~\cite{dhamo2024headgas} extend 3D Gaussians with per-Gaussian basis of latent features to control expressions. NPGA~\cite{giebenhain2024npga} introduces dynamic modules to deform 3D Gaussians and a detail network to generate fine-grained details.  On the other hand, GaussianAvatars~\cite{Qian_2024_CVPR} and SplattingAvatar~\cite{shao2024splattingavatar} built a consistent correspondence between 3D Gaussians and mesh triangles explicitly. In this paper, we use representations corresponding to 3DGS, and our methods utilize GaussianAvatars as the 3D representations in our framework.

\subsection{Image Editing}
To satisfy customized manipulation to a given image, many methods are proposed for image editing using textual instructions. Stable-Diffusion~\cite{rombach2021highresolution} edits specific regions by masking and prompting. DreamBooth~\cite{ruiz2023dreambooth} fine-tunes SD on 3–5 images of a subject to generate personalized edits. ControlNet~\cite{zhang2023adding} adds spatial conditioning to diffusion models via parallel residual connections, Enabling precise structural edits. Prompt-to-Prompt (P2P)~\cite{hertz2022prompt} manipulate cross-attention maps between source and target prompts to guide edits. Uni-ControlNet\cite{zhao2023uni} unifies adapters for global/local control. OmniEdit~\cite{wei2024omniedit} utilize Multimodal large language model (MLLM) to guide image editing. FreeEdit~\cite{he2024freeedit} supports mask-free reference editing by extracting multi-level features via U-Net and injecting them into denoising networks. MIGE~\cite{tian2025mige} proposes a unified multimodal editing framework, which combines CLIP semantic features and VAE visual tokens, processed by LLMs for cross-attention guidance in diffusion. An essential task in image editing is Makeup Transfer, where textual instructions are insufficient to describe the facial makeup accurately. Early image makeup transfer methods\cite{zhu2017unpaired, chang2018pairedcyclegan, li2018beautygan, jiang2020psgan, deng2021spatially, nguyen2021lipstick, xiang2022ramgan, gu2019ladn, liu2021psgan++, wan2022facial, yan2023beautyrec, sun2023ssat, kips2020gan, hu2022protecting, yang2022elegant} first utilize facial landmark extraction and detection to preprocess the face image. Then neural networks are employed to transfer various makeup styles. Methods based on two optimization methods,~\ie, Generative Adversarial Networks(GANs)~\cite{10.5555/2969033.2969125} and Diffusion Model~\cite{10.5555/3495724.3496298}.GAN-based methods have long been utilized in the makeup transfer task. Beauty-GAN~\cite{li2018beautygan} relies on pixel-level Histogram Matching and employs several loss functions to train its primary network. PSGAN~\cite{jiang2020psgan} focuses on transferring makeup between images exhibiting different facial expressions, specifically targeting designated facial areas. CPM~\cite{nguyen2021lipstick} incorporates patterns into the makeup transfer process to transcend basic color transfer. SCGAN~\cite{deng2021spatially} utilizes a part-specific style encoder to differentiate makeup styles for various components. Lastly, RamGAN\cite{xiang2022ramgan} aims to maintain consistency in makeup applications by integrating a region-aware morphing module. Recently, diffusion-based methods have demonstrated their capability in real-world makeup transfer. Stable-Makeup~\cite{zhang2024stablemakeuprealworldmakeuptransfer}  is based on a diffusion framework with multiple controls. It utilizes a Detail-Preserving Makeup Encoder to extract the makeup details, Content and Structural Control Modules to maintain the avatar's identity and Makeup Cross-attention Layers to align the features of the identity embeddings and the makeup embeddings. In this paper, we lift a pretrained Stable-Makup model to 3D avatars to enable 3D makeup transfer.

\begin{figure}[t]
  \centering
  \includegraphics[width=1\linewidth,]{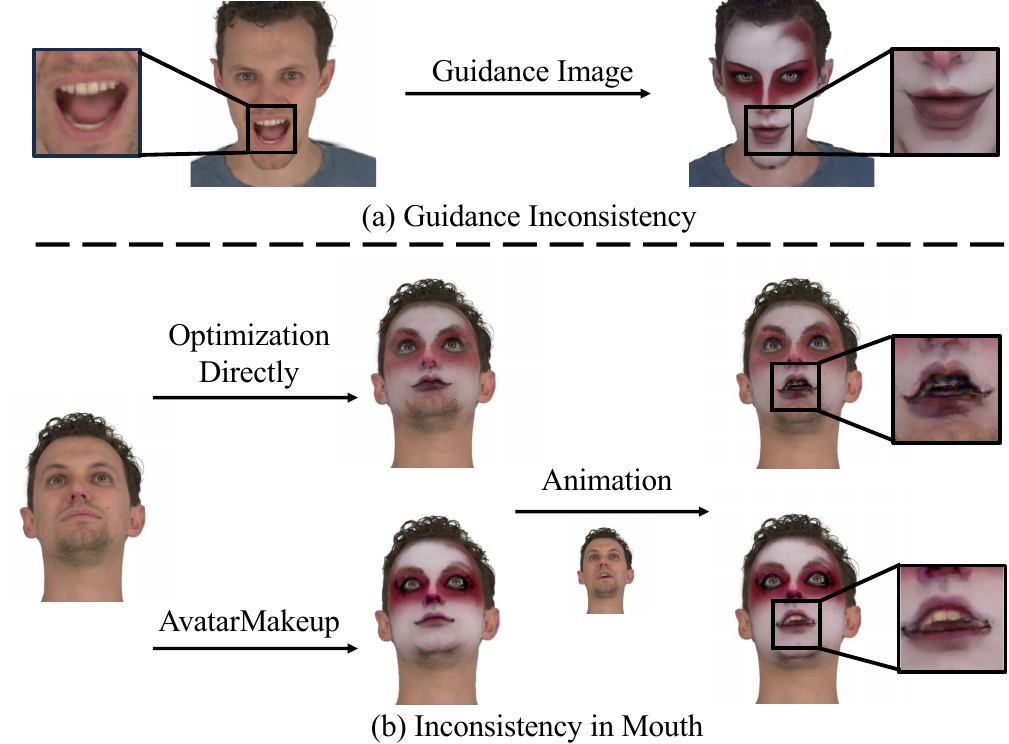}
  \caption{Illustration of the inconsistency during optimization. (a) shows that the mouth is deformed in the guidance image, which is generated by Stable-Makeup. Therefore, directly using these guidance images to optimize the avatars will blur the makeup details. In (b), when optimizing the avatars directly, the teeth' identity will be destroyed during animation. On the contrary, our method adds two proposed strategies and preserves the teeth' identity effectively.}

  \label{fig:motivation}
\end{figure}

\section{Preliminary}
\label{sec:preliminary}
\subsection{GaussianAvatars}
Our makeup model,~\ie, AvatarMakeup, is developed based on 3D models of characters constructed by GaussianAvatars~\cite{Qian_2024_CVPR}. GaussianAvatars employs 3D Gaussian Splatting~\cite{kerbl20233d} as representation to produce high-fidelity human faces. Since the original 3D Gaussian Splatting (3DGS) models are static, GaussianAvatars integrates 3D Gaussian splats with the FLAME~\cite{FLAME:SiggraphAsia2017} mesh by binding Gaussian kernels to mesh triangles, enabling dynamic expressions and movements.
Concretely, a kernel of 3D Gaussian splatting is represented as $\langle \boldsymbol{\mu}, \boldsymbol{s}, \boldsymbol{q}, \boldsymbol{r} \rangle$, where $\boldsymbol{\mu} \in \mathbb{R}^{3}$ denotes the position vector, $\boldsymbol{s} \in \mathbb{R}^{3}$ is the scaling vector, $\boldsymbol{q} \in \mathbb{R}^{4}$ (corrected dimension for quaternion) represents the quaternion, and $\boldsymbol{r} \in \mathbb{R}^{3 \times 3}$ corresponds to the rotation matrix.  As for a FLAME mesh triangle, let $\boldsymbol{T}$ be the mean position of the triangle vertices, a rotation matrix $\boldsymbol{R}$ describes the orientation of the triangle, and a scalar $\boldsymbol{k}$ by the mean length of one of the edges and its perpendicular to denote the scales of the triangle. 
According to the relative position of $\boldsymbol{\mu}$ and triangles, GaussianAvatars bind every gaussian kernel to the nearest triangle. 
When the target face is rigged to another expression, the position of the kernel is updated following the movement of the bound triangle following Eq.~\eqref{eq:globel_rotation}, ~\eqref{eq:global_position} and ~\eqref{eq:global_scaling}.
\begin{align}
    \bm{r}' &= \bm{R} \bm{r},\label{eq:globel_rotation}\\
    \bm{\mu}' &= k \bm{R} \bm{\mu} + \bm{T}, \label{eq:global_position} \\
    \bm{s}' &= k \bm{s} \label{eq:global_scaling}
\end{align}

The rendering process is a standard 3DGS rendering, which computes the color of a pixel by blending all Gaussians overlapping the pixel following Eq.~\eqref{eq:rendering}.
\begin{equation}
\boldsymbol{C}=\sum_{i=1} \boldsymbol{c}_{i} \alpha_{i}^{\prime} \prod_{j=1}^{i-1}\left(1-\alpha_{j}^{\prime}\right)\label{eq:rendering}
\end{equation}

\subsection{Stable-Makeup}
In this paper, we use Stable-Makeup to generate makeup guidance to supervise the target avatars. Stable-Makeup~\cite{zhang2024stablemakeuprealworldmakeuptransfer} introduces a diffusion-based approach for robust real-world makeup transfer. At its core, Stable-Makeup leverages a pre-trained diffusion model and incorporates three key innovations to enable precise makeup transfer while preserving the identity of the original avatars. First, given a reference makeup image $I_m$ and an original image of the target avatar $I_t$, Stable-Makeup extracts multi-scale makeup details from $I_m$ using a \textit{Detail-Preserving Makeup Encoder}. This encoder employs a pre-trained CLIP~\cite{radford2021learningtransferablevisualmodels} model to extract features from multiple layers, which are concatenated and processed by self-attention to capture local and global makeup features, preserving fine-grained makeup details. Second, Stable-Makeup proposes \textit{Makeup Cross-Attention Layers} to align the makeup embeddings with the source image's facial structure. Third, Stable-Makeup employs \textit{Content and Structural Control Modules} based on ControlNet~\cite{zhang2023adding} to maintain the $I_t$’s identity. The content encoder preserves pixel-level consistency of $I_t$, while the structural encoder introduces facial structure control using dense lines derived from facial landmarks. These modules are formulated as
\begin{equation}
y_c = \mathcal{F}(x; \Theta) + \mathcal{Z}\left( \mathcal{F}\left( x + \mathcal{Z}(c; \Theta_{z1}); \Theta_c \right); \Theta_{z2} \right), 
\end{equation}
where $\mathcal{F}$ is the U-Net, $\Theta$ are frozen weights, $\Theta_c$ are trainable ControlNet weights, and $\mathcal{Z}$ denotes zero-convolution layers. This design ensures that the generated image $I_t$ retains the identity of the source.
During training, the loss function of Stable-Makeup extends the standard diffusion objective:
\begin{equation}
    \mathcal{L}_{SM} = \mathbb{E}_{x_0, t, \epsilon} \left[ \left\| \epsilon - \epsilon_\theta \left( x_t, t, c_i, c_e, c_m \right) \right\|_2^2 \right], 
\end{equation}
where $c_i, c_e, c_m$ are content, structural, and makeup conditioning inputs, respectively. This forces the model to ensure the identity of $I_t$ and makeup patterns of  $I_m$.  

\section{The Proposed Method}
\label{sec::method}

In this section, we present AvatarMakeup for transferring the makeup patterns from an individual's face to 3D avatars. Since previous methods like GaussianEditor use textual instructions for editing, we conduct experiments using textual instructions to guide the makeup transfer and find that it results in low-quality effects. The comparison results are shown in Sec.~\ref{subsec:Comparison}. On the contrary, we believe that transferring makeup from a single reference image of any individual provides more rich and precise makeup details. Given the reference image, we lift a diffusion-based model,~\ie, Stable-Makeup, to 3D space. Recent methods,~\eg, Score Distillation Sampling(SDS)~\cite{poole2022dreamfusion} and DreamLCM~\cite{10.1145/3664647.3680709} provide a feasible way to achieve this, which utilizes the guidance images generated by Stable-Makeup. However, the images generated by the diffusion models are inconsistent with the target avatars, resulting in the artifacts shown in Fig.~\ref{fig:motivation}(a). Innovatively, we adopt a \textit{coarse-to-fine} idea to first apply base makeup to the avatars and then enhance the details.
The coarse stage employs a global UV map to ensure consistent makeup effects, effectively avoiding artifacts typically caused by diffusion models. 
The overall structure of AvatarMakeup is illustrated in Fig~\ref{fig:pipeline}. 
The Base Makeup stage, illustrated in Fig~\ref{fig:pipeline}\textcolor{red}{(1)}, takes as input an animatable avatar generated by GaussianAvatars~\cite{Qian_2024_CVPR}.
We propose a Coherent Duplication method in Sec~\ref{sec:4.2} to generate highly consistent base makeup. With the Coherent Duplication stage, the avatars' makeup is consistent across multiple viewpoints and expressions. The refinement stage is shown in Fig~\ref{fig:pipeline}\textcolor{red}{(2)}. 
Input the optimized avatars from the coarse stage, we integrate a Refinement Module to generate refined guidance with richer makeup details in Sec.~\ref{sec:4.3}. 
\subsection{Coherent Duplication}
\label{sec:4.2}
In this subsection, we aim to utilize Stable-Makeup's advanced image makeup transfer ability and handle the inconsistency issue in previous methods. Previous methods such as DreamFusion ~\cite{poole2022dreamfusion} use a differentiable renderer to render images of target avatars. They optimize avatars based on the discrepancy between rendered images and guidance images which are generated by image generation methods. However, the guidance images generated by Stable-Makeup differ from the original avatars and other genereated guidance images. Therefore, directly using the guidance to optimize avatars leads to inconsistency. As shown in Fig~\ref{fig:motivation}(a) and (b), the guidance images generated by Stable-Makeup show a misaligned facial contour with the original avatar image and missing teeth. The misalignment not only inevitably introduces noisy artifacts but also destroys the integrity of the avatar's inner structure, e.g., teeth, tongue, during optimization. Besides, the inconsistency between the guidance images causes over-smooth makeup. Conventional methods utilize a UV map to record the texture of a mesh-based head. Despite the fact that the UV map falls short in rendering high-detailed textures, the UV map retains consistent textures, avoiding the above issues. Inspired by this, we design a two-stage training strategy. In the coarse stage, we generate base makeup using a proposed \textit{Coherent Duplication (CD)} module, which utilizes a global UV map to maintain the consistency of the target appearance.
 
Particularly, given rendered facial images of 3DGS $I$ along with a reference makeup image, we first use the Stable-Makeup network $\mathcal{F}_\theta$, parameterized by $\theta$, to generate guidance images $I_\theta$. We experimentally find that Stable-Makeup generates detailed makeup images and the makeup aligns well with the facial region when target avatars are under canonical expressions. We then render images after driving the avatars to canonical expressions and utilize the rendered images to generate coherent guidance images. Notably, using a single view guidance image to generate the UV map causes defects due to facial occlusion. We fill the global UV map by accumulating $N$-view guidance images. We denote the guidance images with canonical expression as $I^{cano}_{\theta}$. Secondly, we map each pixel $(H,W)$ of $I^{cano}_{\theta}(H,W)$ to the pixels on the UV map $(h,w)$, where $(h,w)$ and $(H,W)$ denote the pixel position. Here, we use a mesh renderer to directly render the mapping images, denoted as $I_{map}$. Given $I^{cano}_{\theta}$ and $I_{map}$, we then optimize the UV map formulated following Eq.~\eqref{map}.
\begin{align}
\label{map}
\resizebox{0.9\hsize}{!}{$I_{UV}(h,w) = \sum_{i=1}^N \frac{1}{\mid S_i\mid}\sum_{H,W}I_{\theta}^{cano_i}(H,W), where (H,W)\in S_i)$},
\end{align} 
where $I(h,w)$ represents the RGB values of each pixel, and $S_i=\{(H,W)\mid I_{map}(H,W)=(h,w)\}$. Since the UV map remains constant, it provides global makeup details. By querying the UV map, we then render coherent guidance images $I_{UV}$ across multiple viewpoints and expressions. In practice, we can easily obtain $I_{UV}$ using the mesh renderer. We use the coherent guidance images to optimize the avatar, resulting in highly consistent makeup effects. However, the UV map has limited resolution, which leads to low-quality makeup effects. Besides, the details in the eyes and the hair region are blurred. Therefore, we employ several strategies to enhance facial details in Section~\ref{sec:4.3}.

Overall, the coarse stage training utilizes Coherent Duplication module to generate base makeup for the avatars, ensuring both (1) makeup consistency during animation and (2) provision of coherent priors for the subsequent refinement module.

\subsection{Detail Refinement}
\label{sec:4.3}
Since the base makeup generated by Coherent Duplication exhibits spatial consistency but suffers from limited visual quality, we propose a \textit{Detail Refinement (DR)} module in the refinement stage training to enhance makeup details while maintaining geometric coherence. This module utilizes the base makeup as structural priors to guide the refinement process. The core idea of the proposed module is to leverage the priors to preserve consistency and forward Stable-Makeup for generating refined makeup guidance. Formally, let $\hat{I} $ denote the base makeup rendered from coarsely optimized avatars, and $I_{m}$ represent the reference makeup image. Stable-Makeup proceeds with the diffusion process to obtain the refined guidance images $\hat{I}_{\theta}$. In the diffusion process, we integrate the refinement module by injecting noise at small timestamps $t$. Crucially, $\hat{I}_{\theta}$ preserves structural consistency while significantly enhancing makeup details. Finally, we optimize the avatars using these refined guidance images, achieving high-fidelity makeup avatars.

During optimization, we assume that the 3D Gaussians are optimally distributed on the FLAME mesh to express all kinds of poses and expressions. Consequently, we freeze the Gaussian attributes $\{\mathbf{x}, \mathbf{r}, \mathbf{s}\}$,~\ie, position, rotation, scale, and only optimize the parameters of the feature $\mathbf{f}$ and opacity $\alpha$ . This preserves the avatar's geometric structure while eliminating the need for adaptive density control~\cite{kerbl20233d}. Moreover, the coherent guidance images generated in the Coherent Duplication method and these
sections both exhibit blurred facial details in two aspects: 1) Due to the rendering process of 3D Gaussians which is accumulating multiple 3D gaussians, the facial color in the same position may vary across different viewpoints and expressions. 2) Directly optimizing avatars destroys facial details in non-makeup region, disadvantages in preserving the identity of the avatars,~\eg, the details of the teeth are destroyed during optimization in Fig.~\ref{fig:motivation}(b). We propose two strategies to enhance facial details. For the \textit{first} issue,  we generate guidance images covering multiple viewpoints and expressions. For the \textit{second} issue, we employ a face-parsing model~\cite{10.1007/s11263-021-01515-2} to create precise masks that isolate the makeup regions for optimization. We further introduce restirction loss to supervise non-makeup region of target avatars with the identity-preserving images rendered from the original avatars. For each rendered image $I_r$, we obtain the corresponding guidance image $I_G$, mask image $M$ and identity image $I_{ID}$ under consistent viewpoint and expression conditions. In particular, in Coherent Duplication, $I_G$=$I_{UV}$, while in Detail Refinement, $I_r$=$\hat{I}$ and $I_G$=$\hat{I}_{\theta}$. Consequently, in both CD and DR modules, we supervised the makeup details with $\mathcal{L}_1$ loss and LPIPS loss in Eq.~\eqref{makeup_loss}.
\begin{align}
\label{makeup_loss}
\scalebox{0.9}{$\mathcal{L}_{\text{makeup}} = \mathcal{L}_1(M\odot I_G, M\odot I_r) + \mathcal{L}_{\text{LPIPS}}(M\odot I_G, M\odot I_r).$}
\end{align} 
We then employ the restriction loss,~\ie, Eq.~\eqref{restriction_loss}, to preserve the identity,~\ie, the non-makeup region.
\begin{equation}
\label{restriction_loss}
\mathcal{L}_{\text {Res}} = \mathcal{L}_1((1-M)\odot I_{ID}, (1-M)\odot I_r). 
\end{equation}
The total loss is in Eq.~\eqref{total_loss}.
\begin{equation}\label{total_loss}
\mathcal{L} = \lambda_1 \mathcal{L}_{makeup} + \lambda_2 \mathcal{L}_{\text{Res}},
\end{equation} where $\lambda_1$ and $\lambda_2$ are loss weights. 
\begin{figure}[t]
  \centering
  \includegraphics[width=1\linewidth,]{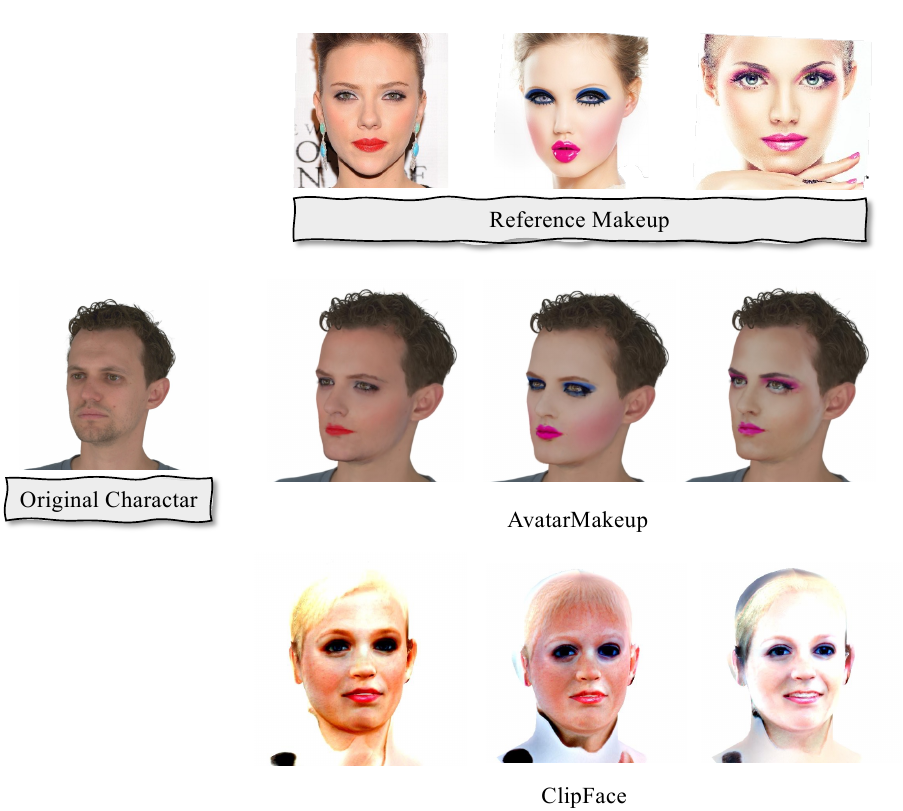}
  \caption{Qualitative comparision between our methods and ClipFace~\cite{aneja2023clipface}. On the one hand, we can see that our methods successfully tranfer fine-grained makeup details to the target avatars, while ClipFace totally fail  to maintain the identity and makeup information. On the other hand, our methods preserves the identity better than ClipFace. The ClipFace generates characters look like the avatars in the reference image, while our method preserve the identity of the target avatar.}
  \label{fig:clipface}
\end{figure}

\begin{table*}[t]
\centering
\begin{subtable}[t]{\textwidth}
\centering
\begin{tabularx}{\linewidth}{l|XXXX|XXXX}
\toprule
&\multicolumn{4}{c|}{multi-view DINO-I↑} &\multicolumn{4}{c}{animation DINO-I↑}
\\ 
&$0^{\circ}$ & $45^{\circ}$ &$-45^{\circ}$ &average &$0^{\circ}$ & $45^{\circ}$ &$-45^{\circ}$ &average
\\ \hline
ClipFace \cite{aneja2023clipface} &0.381 &0.339 &0.338 &0.353 & 0.363 & 0.316 & 0.332 & 0.337  \\
Ours      & \textbf{0.726} & \textbf{0.620} & \textbf{0.626} &\textbf{0.656} &\textbf{0.695} &\textbf{0.590} &\textbf{0.596} &\textbf{0.627}\\ \bottomrule

\end{tabularx}

\caption{Multi-view DINO-I metric and Animation DINO-I metric.}
\end{subtable}

\begin{subtable}[t]{\textwidth}
\centering
\begin{tabularx}{\linewidth}{p{2cm} | *{4}{>{\centering\arraybackslash}X|} >{\centering\arraybackslash}X }
\toprule
 & FID↓ & KID↓ & GPT-4o(MS)↑ &GPT-4o(MQ)↑ &GPT-4o(IP)↑\\ \hline
ClipFace &160.6  & 0.155 & 3.64 &2.38 &3.48 \\
Ours & \textbf{152.0} & \textbf{0.130} &\textbf{4.04} &\textbf{3.78} &\textbf{4.98}  \\ 
\bottomrule
\end{tabularx}
\caption{FID, KID and AIME metric.}
\end{subtable}
\caption{Quantitative comparison with the baseline. We can see that AvatarMakeup surpassed the existing baselines in numerical results, demonstrating the superiority of our methods in makeup quality. }
\label{tab:sota}
\end{table*}

\begin{table*}[t]
\centering
\begin{subtable}[t]{\textwidth}
\centering
\begin{tabularx}{\textwidth}{p{3.7cm}|XXXX|XXXX}
\toprule
&\multicolumn{4}{c|}{DINO-I↑}     & \multicolumn{4}{c}{CLIP-I↑} 
\\ 
&$0^{\circ}$ & $45^{\circ}$ &$-45^{\circ}$ &average & $0^{\circ}$ &$45^{\circ}$ &$-45^{\circ}$ &average 
\\ \hline
Vanilla & 0.698 & 0.585 & 0.591 & 0.625 &0.656 &0.608 &0.617 &0.627   \\
\hline

w/o Coherent Duplication &0.700 &0.568 &0.572 &0.613& 0.644 & 0.606 & 0.592 & 0.614  \\
w/o Detail Refinement &0.692 &0.582 &0.579 &0.618 &0.634 &0.595 &0.588 &0.606\\ 
full      &\textbf{0.726} &\textbf{0.620} &\textbf {0.626} &\textbf{0.656} &\textbf{0.678} &\textbf{0.619} &\textbf{0.626} &\textbf{0.641} \\ \bottomrule
\end{tabularx}
\caption{Multi-view Makeup Transfer. }
\end{subtable}
\begin{subtable}[t]{\textwidth}
\centering
\begin{tabularx}{\textwidth}{p{3.7cm}|XXXX|XXXX}
\toprule

\multicolumn{1}{c|}{}  &\multicolumn{4}{c|}{DINO-I↑ }     & \multicolumn{4}{c}{CLIP-I↑} 
\\  &$0^{\circ}$ & $45^{\circ}$ &$-45^{\circ}$ &average & $0^{\circ}$ &$45^{\circ}$ &$-45^{\circ}$ &average 
\\ \hline
Vanilla &0.671 &0.561 &0.569 &0.600 & 0.644 & 0.612 & 0.602 & 0.619  \\ \hline
w/o Coherent Duplication &0.672 &0.548 &0.554 &0.591 &0.640 & 0.606 & 0.591 & 0.612 \\
w/o Detail Refinement &0.658 &0.553 &0.550 &0.587 &0.625 &0.594 &0.579 &0.600\\ 
full &\textbf{0.695} &\textbf{0.590} &\textbf{0.596} &\textbf{0.627} &\textbf {0.664} &\textbf {0.621} &\textbf {0.610} &\textbf {0.632} \\ \bottomrule
\end{tabularx}
\caption{Animation Makeup Transfer.}
\end{subtable}

\caption{We conducted ablation experiments on each module. The results demonstrate that each module contributes effectively to the overall makeup effects.}
\label{tab:ablation}
\end{table*}

\vspace{-1em}\section{Experiments}
\label{sec:experiment}
\subsection{Implementation}
The proposed AvatarMakeup method leverages well-constructed gaussian avatars from GaussianAvatars~\cite{Qian_2024_CVPR}. StableMakeup~\cite{zhang2024stablemakeuprealworldmakeuptransfer} serves as the guidance model for the image makeup transfer process.
In the base makeup stage, the resolution of the UV map is set to 256×256. We use 16 different-view fuidance images under canonical expression to fill the UV map. For the Detail Refinement module, we linearly sample timestamps t$\in [20,400]$ for the forward diffusion process. In both stages, we render images at a resolution of 512×512 to align with the standard input requirements of Stable-Makeup and the face-parsing model~\cite{10.1007/s11263-021-01515-2}. When using Stable-Makeup to generate guidance, we configure the inference steps to 50 in the base makeup stage to generate high-quality makeup and 5 in the refinement stage to execute fast refinement. We obtain guidance images with 5,000 different expressions and viewpoints in the base makeup stage and 3,000 in the refinement stage to maintain high-quality makeup results during animation. To enable sufficient training, the overall transfer process consists of 13,000 iterations, with 10,000 steps allocated to the first stage and the remaining 3,000 steps dedicated to the refinement stage. During optimization, we set the loss weights $\lambda_1=\lambda_2=10.0$ and use the Adam~\cite{kingma2014adam} optimizer for gradient descent. We set $sh=0$ in practice and the learning rate to $1e-3$ to optimize the opacity and feature properties of 3D gaussians. 

\subsection{Evaluation Settings}
\textbf{Datasets.} We utilize two datasets for evaluation,~\ie, \textbf{NeRSemble~\cite{kirschstein2023nersemble}} dataset and  \textbf{LADN~\cite{gu2019ladnlocaladversarialdisentangling}} dataset to obtain reconstructed 3D avatars and reference makeup images, respectively.
\begin{itemize}
\item \textbf{NeRSemble~\cite{kirschstein2023nersemble}} records 11 video sequences for each avatar. Each frame of the sequences contains 16 camera views surrounding the avatar. The first 10 sequences are obtained by asking the participants to perform the expression following the instructions. Particularly, the $11^{th}$ video sequence is a free-play sequence. We sample expressions in the first 10 video sequences for training and the $11^{th}$ sequence for evaluation. During evaluation, we select 9 avatars from the dataset and reconstruct using GaussianAvatars~\cite{qian2024gaussianavatars} methods.
\item \textbf{LADN~\cite{gu2019ladnlocaladversarialdisentangling}} contains real-world makeup images containing simple and complicated makeup patterns. We randomly select 50 images as reference makeup images for quantitative comparison.\\
\end{itemize}
\begin{figure*}[!t]
  \centering
  \includegraphics[width=0.98\linewidth,]{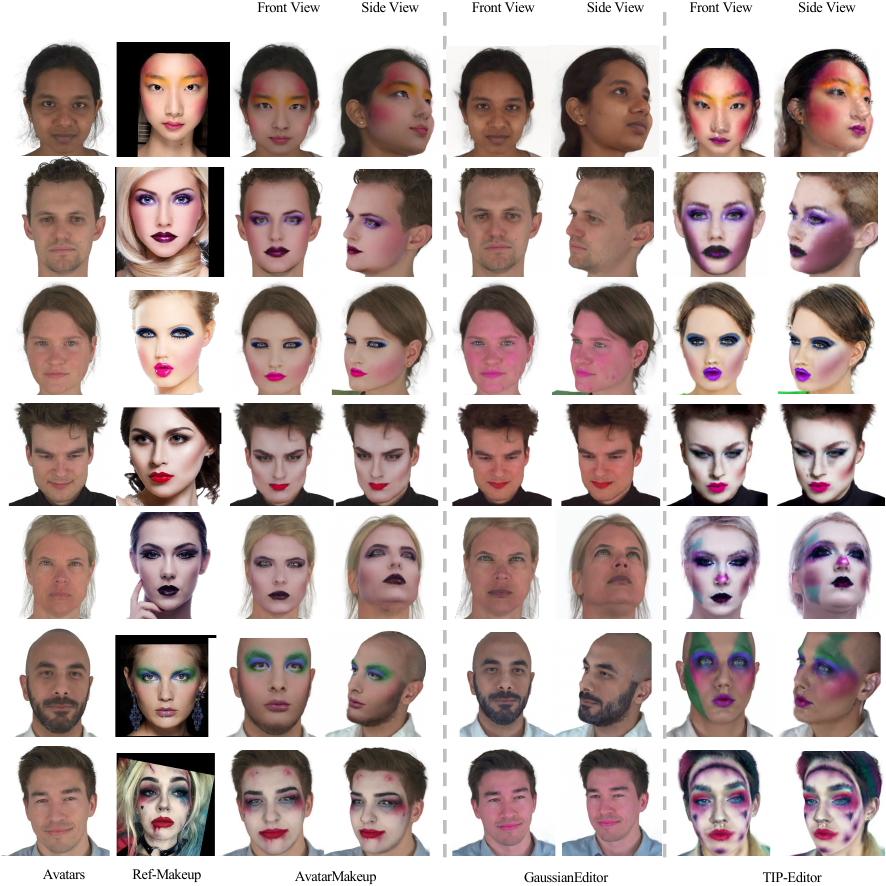}
  \caption{\textbf{Qualitative Comparison.} GaussianEditor~\cite{chen2024gaussianeditor} alters the face color but generates low-quality eye shadow. TIP-Editor~\cite{zhuang2024tip} struggles to preserve the identity of the original avatars while generating incorrect makeup colors, such as the mismatched lips color in the first row and the face color in the second row. In contrast, AvatarMakeup accurately transfers makeup details while preserving the avatar's identity. Besides, AvatarMakeup supports animations, which are not available in the baseline methods.  }
  \label{fig:qualitative}
\end{figure*}

\textbf{Criteria.} 
Since this is the first work to achieve makeup transfer to 3D Gaussian avatars, we adapt evaluation criteria from relevant 3D Gaussian editing and 2D image editing methods,~\eg, , Stable-Makeup~\cite{zhang2024stablemakeuprealworldmakeuptransfer} and ClipFace~\cite{aneja2023clipface}.
Specifically, we use the following metrics to evaluate makeup transfer quality and identity preservation:
\begin{itemize}
    \item \textbf{DINO-I~\cite{oquab2023dinov2}}: It utilizes a DINO backbone to extract dense features and calculates the cosine similarity between the features of the target image and the makeup image.
    \item \textbf{Fr\'{e}chet Inception Distance (FID)~\cite{heusel2017gans}}: It quantifies the similarity between the generated and real image distributions using the Fr\'{e}chet distance in the feature space of a pretrained Inception-v3 network~\cite{szegedy2016rethinking}.
    \item \textbf{Kernel Inception Distance (KID)~\cite{binkowski2018demystifying}}: It measures the squared Maximum Mean Discrepancy (MMD) between feature distributions using an unbiased polynomial kernel.
    \item \textbf{AI-Assisted Makeup Evaluation (AIME)}. This proposed metric leverages advanced Multimodal Large Language Models (MLLMs),~\eg, GPT-4o~\cite{hurst2024gpt}, to provide a nuanced assessment of both makeup transfer quality and identity preservation. Specifically, we concatenate the original rendered image, the reference makeup image, and the makeup-transferred image together in the width dimension into one example. Subsequently, we feed the example to gpt-4o and ask it to score it from 1 to 5 in the following aspects: 1) \textit{makeup similarity} to judge the fidelity of the generated makeup to the reference makeup ; 2) \textit{makeup quality} to evaluate the makeup transfer quality; 3) \textit{identity preservation} to evaluate structural consistency with the original avatars.
\end{itemize}
\begin{figure*}[p]
  \centering
  \includegraphics[width=1\linewidth,]{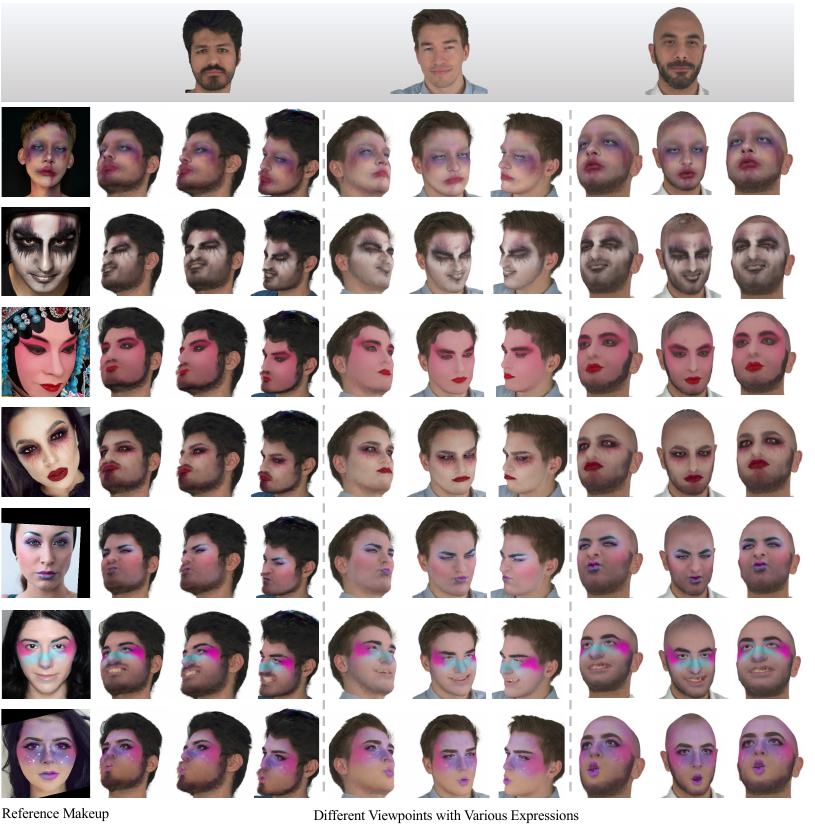}
  \caption{Additional makeup results generated using AvatarMakeup. Given a real-world reference makeup, our methods can transfer the makeup pattern to the target 3D avatars with fine-grained details, while maintaining the original identity. Besides, under animation and multiview condition, the makeup maintains high-quality with negligible artifacts. Zooming in is recommended to observe the high-resolution details. }
  \label{fig:results}
\end{figure*}
For both FID and KID, we calculate the similarity between the reference makeup images and the rendered images from the target avatars.
We conduct experiments to evaluate the quantitative results of 3D makeup transfer under two settings: \textit{Multi-view Makeup Transfer} to evaluate the makeup consistency under multi-view condition, and \textit{Animation Makeup Transfer} to evaluate makeup consistency under both multi-expression and multi-view conditions. For the former, we evaluate the results under canonical expression for each avatar rendered from three specific views, with azimuth angles set to 45°, 0°, and -45°, and the elevation angle fixed at 0°. For the latter, we randomly sample 5 FLAME parameters on the $11^{th}$ video sequence in NeRSemble dataset for each subject. In this case, the facial expressions are randomly sampled from a distribution distinct from the training set, representing novel, unseen expressions during evaluation. For each expression, we render images from the same viewpoints as in the \textit{Multi-view Makeup} configuration.  We conduct qualitative comparisons to demonstrate the high makeup quality of our method. 

\textbf{Baselines.} We evaluate quantitative and qualitative results using different baselines. For quantitative results, we train AvatarMakeup and ClipFace~\cite{aneja2023clipface}. ClipFace generates 3D avatars by combining a StyleGAN-based network and FLAME-based mesh. The method enables avatars editing by minimizing the CLIP loss between the target avatars and the text instructions. Additionally, avatars can be animated by FLAME parameters. To achieve makeup transfer, we first employ GAN inversion to train ClipFace with specific avatars. We then utilize the CLIP loss between the target avatars and the reference makeup images to optimize the avatars. Since the FLAME parameter are constant during optimization, ClipFace can preserve the avatars' geometric structure.

For qualitative evaluation, we choose GaussianEditor~\cite{chen2024gaussianeditor} and TIP-Editor~\cite{zhuang2024tip} as the baseline methods. We do not compare with DGE~\cite{chen2024dge} since the method does not generate reasonable effects in our experiments. Crucially, the baseline methods and our methods use different conditions to control the transferring process. Our method takes the reference makeup images as the condition. GaussianEditor uses textural instructions, and TIP-Editor achieves makeup transfer using both text and reference images as condition. For a fair comparison, we preprocess the baselines before evaluation as follows:
\begin{itemize}
    \item \textbf{GaussianEditor.} Given textual instructions, GaussianEditor edit 3D gaussians using image editing methods such as Instruct Pixel2pixel~\cite{brooks2023instructpix2pix}. Therefore, we use GPT-4o~\cite{openai2024gpt4technicalreport} to generate textual descriptions for the reference makeup. Specifically, for each reference makeup image, we input the image and the prompt "\textit{describe the detailed facial makeup in the image in one sentence}" to gpt-4o. We then use the output sentence by gpt-4o, along with the rendered images of the target avatars achieve to apply GaussianEditor to generate makeup transfer results.
    \item \textbf{TIP-Editor.} TIP-Editor combines textual instructions and image condition to generate both semantic and low-level features, allowing for accurate editing. Given the rendered images denoted as \textit{$<$src$>$} and reference makeup images denoted as \textit{$<$ref$>$}, we integrate the images into the following sentence "\textit{a photo of a $<$src$>$ person with $<$ref$>$ makeup style}" as prompt. We then input the prompt into TIP-Editor to execute makeup transfer.
\end{itemize}

\label{sec:comparison}
\subsection{Comparisons}
\label{subsec:Comparison}
\textbf{Qualitative Results.} The qualitative experiments results are shown in Fig.~\ref{fig:qualitative}. We compare our methods with GaussianEditor and TIP-Editor by displaying makeup effects in the front view and a randomly sampled view. Our method shows superiority in two aspects. On the one hand, our results exhibit high-quality makeup transfer results. We can see that in the third row, GaussianEditor does not transfer the eye shadow and alters the face color, and TIP-Editor generates incorrect lip color. In the fifth row, GaussianEditor generates very light makeup, and TIP-Editor generates noisy artifacts, destroying the makeup pattern. In contrast, AvatarMakeup generates delicate makeup without artifacts. On the other hand, our results maintain the avatar's identity. For example, all the examples show that TIP-Editor tends to generate the identity of the reference makeup. AvatarMakeup preserves the identity of the original avatars. In the comparison between AvatarMakeup and ClipFace shown in Fig.~\ref{fig:clipface}, we can see that ClipFace diffuses makeup to all facial regions while our methods accurately align the makeup with specific facial regions. Moreover, GaussianEditor and TIP-Editor can handle only static avatars. We further display more generated results under multiview condition and animation conditions, shown in Fig.~\ref{fig:results}.\\ 
\textbf{Quantitative Results.} We conduct quantitative experiments by calculating the four metrics comparing our methods and ClipFace~\cite{aneja2023clipface}. The results are shown in Tab~\ref{tab:sota}. We can see that AvatarMakeup outperforms ClipFace in the DINO-I metric. Remarkably, AvatarMakeup achieves 65.6$\%$ in DINO-I metric, which is a 30.3$\%$ huge improvement than ClipFace, indicating that AvatarMakeup generates high-fidelity makeup to reference makeup. Besides, AvatarMakeup scores lower FID(152.0) and KID(0.130) than ClipFace. This reflects that our method generates more realistic makeup images close to real-world images. Beyond traditional comparisons using visual metrics, we further evaluate our AIME metric to judge makeup transfer with human preference. The results show that in all three aspects, AvatarMakeup gets higher scores than ClipFace. Notably, Avatar Makeup has 3.78 MQ quality, compared to 2.38 in ClipFace. The improvement demonstrates that AvatarMakeup generates high-quality makeup effects. Overall, the quantitative results demonstrate that AvatarMakeup has superior makeup transfer quality than state-of-the-art methods.\vspace{-0.5em}
\subsection{Ablation Study} 
We first explore the effect of coherent duplicate modules by removing the module while keeping the rest of the experimental setup. Secondly, we explore the effect of the coarse stage. Concretely, we evaluate the makeup on the avatars optimized without the refinement stage. We design a vanilla version that directly optimizes the avatars using guidance images generated by Stable-Makeup. Table~\ref{tab:ablation} shows the ablation results. The results show lower CLIP-I score(-3.4$\%$ in Multi-view Makeup Transfer(MT) and -2.4$\%$ in Animation MT) and DINO-I score(-2.6$\%$ in Multi-view MT and -2.3$\%$ in Animation MT) after deleting the Coherent Duplication module. The numerical decrease exists when deleting the Detail Refinement module or in the Vanilla version, which demonstrates that every module is effective in generating consistent and high-quality makeup effects.

\section{Conclusion}
We proposed AvatarMakeup, a 3D makeup transfer method that ensures consistent appearance during animations, preserves identity, and enables fine detail control. By combining a pretrained diffusion model with a coarse-to-fine strategy, our approach uses Coherent Duplication to achieve multiview and dynamic consistency and a Refinement Module for enhanced makeup quality. Experimental results demonstrate that AvatarMakeup outperforms existing methods in both quality and consistency, providing a robust solution for realistic 3D avatar customization.
\bibliographystyle{IEEEtran}
\bibliography{Ref}

\end{document}